\documentclass[journal]{IEEEtran}

\usepackage{cite}
\usepackage{graphicx}
\usepackage{amsmath}
\usepackage{algorithm}
\usepackage{algpseudocode} 
\usepackage{varwidth}
\usepackage{ifpdf}

\usepackage{subcaption}
\graphicspath{{figs/}}

\begin{document}

\title{Deep Reinforcement Learning-Based Product Recommender for Online Advertising}

\author{Milad~Vaali~Esfahaani, 
        Yanbo~Xue, 
        and~Peyman~Setoodeh
\thanks{M. Vaali Esfahaani and P. Setoodeh are with the School
of Electrical and Computer Engineering, Shiraz University, Shiraz,
Iran (e-mail: m.vaali@shirazu.ac.ir; psetoodeh@shirazu.ac.ir).}
\thanks{Y. Xue is with the Career Science Lab, Beijing, China, and also with the Department of Control Engineering, Northeastern University 
Qinhuangdao, China (e-mail: yxue@careersciencelab.com).}
}

\maketitle

\begin{abstract}
In online advertising, recommender systems try to propose items from a list of products to potential customers according to their interests. Such systems have been increasingly deployed in E-commerce due to the rapid growth of information technology and availability of large datasets. The ever-increasing progress in the field of artificial intelligence has provided powerful tools for dealing with such real-life problems. Deep reinforcement learning (RL) that deploys deep neural networks as universal function approximators can be viewed as a valid approach for design and implementation of recommender systems. This paper provides a comparative study between value-based and policy-based deep RL algorithms for designing recommender systems for online advertising. The RecoGym environment is adopted for training these RL-based recommender systems, where the long short term memory (LSTM) is deployed to build value and policy networks in these two approaches, respectively. LSTM is used to take account of the key role that order plays in the sequence of item observations by users. The designed recommender systems aim at maximising the click-through rate (CTR) for the recommended items. Finally, guidelines are provided for choosing proper RL algorithms for different scenarios that the recommender system is expected to handle. 
\end{abstract}


\section{Introduction}\label{sec:introduction}

Recent advances in information technology and computer networks as part and parcel of the fourth industrial revolution have led to generation and distribution of large amounts of data on a daily basis. Hence, handling big data has become a daily-life challenge for different communities. This challenge stems from the fact that users must be able to effectively use all available resources. Therefore, they will face massive amounts of data, and in order to consciously choose what to be aware of or what to attend to calls for a comprehensive understanding of the content of all options \cite{batmaz2019review}. Such a challenge can be best addressed by deploying software-defined recommender systems. For instance, in e-commerce, recommender systems provide selected useful and valuable information to costumers on one hand and increase sales by targeted advertising on the other hand.

Different techniques and models can be used to implement a recommender system. Recently, machine learning algorithms - a branch of artificial intelligence - have been used for this purpose \cite{lampropoulos2015machine}. These algorithms are divided into three general categories: supervised, unsupervised, and reinforcement learning \cite{portugal2018use}. In supervised learning, training data includes both input and target vectors, so that the desired output can be used in the training process. On the contrary, in unsupervised learning, the dataset does not include target vectors; therefore, such algorithms usually try to identify similarities between samples. In RL algorithms, interaction with the environment allows for finding a suitable action that increases the reward signal \cite{bishop2006pattern}.  

There are three major categories of RL algorithms: actor-only, critic-only, and actor-critic \cite{Grondman2012}. Actor-only algorithms such as policy gradient (PG) method learn a policy function, which determines the action to be taken in a certain state, critic-only algorithms such as Q-learning learn a value function that provides an estimate of the expected collected reward over the control horizon, and actor-critic algorithms learn both a policy function and a value function that evaluates the policy. In the context of deep RL, deep neural networks, which are universal function approximators, are trained to approximate the optimal policy function, value function, or both. In this framework, the corresponding neural networks are called policy and value networks. 

Deploying RL for design and implementation of recommender systems has two main advantages compared to alternative approaches. First, for finding an optimal strategy, RL takes account of the interacting nature of the recommendation process as well as probable changes in customers' priorities over time. These systems consider dynamics of changing tastes and personal preferences, hence, they are able to behave more intelligently in recommending goods. Second, RL-based systems gain an optimal strategy by maximizing the expected accumulated reward over time. Thus, the system with small immediate rewards identifies an item to offer but provides an immense contribution to rewards for future recommendation \cite{zhao2017deep}. Substantial work has been done on using RL in recommender systems aimed at different applications. Most of the reported work in the literature have used variations of either the deep Q-network (DQN) or the PG algorithm. 

Collaborative filtering was used in \cite{art1},   which was then, reformulated as the k-arm bandit problem. This work assumed interdependence between items, which was previously ignored. In the proposed model, interdependent items were considered as clusters of arms, where the arms in a cluster showed the invisible similarities of the items. Using DQN, a system for news recommendation was proposed in \cite{zheng2018drn}. In the proposed recommender system, user's return pattern provides additional information to the recommender system, which is complementary to the click/no-click feedback from the user. A video recommender was proposed in \cite{tripathi2018role}, which uses a hybrid model combined with DQN. Recommendations are based on the content of the video and the feeling it induces. A movie recommender was presented in \cite{zhao2018deep} that uses double DQN to address the overestimation issue. In \cite{chen2018stabilizing}, a robust version of the DQN algorithm was suggested for tip recommendation. In \cite{zhao2019deep}, DQN was used for optimal advertisement and finding optimal location to interpolate an ad in a recommendation list. A social attentive version of DQN was used in \cite{lei2019social}, which benefits from preferences of both users and their social neighbours for estimating action-values. In \cite{pmlr-v97-chen19f}, it was proposed to develop a generative adversarial network (GAN) as a model for user behaviour, which can play the role of the environment for a DQN-based recommender system. 

A conversational recommender system was proposed in \cite{sun2018conversational}, which integrates a dialog system and a recommender system. Using a policy network, the presented system provides personalized recommendations based on the past collective information and the current conversation. The PG algorithm was deployed in \cite{pan2019policy} for contextual recommendations by relaxing some limiting assumptions such as simple reward function and the static environment in the contextual bandit method. A scaled-up version of the REINFORCE algorithm was used in \cite{chen2019top} to deal with a large action space and recommend the top k best items instead of just the best one. Deploying the GAN architecture, in \cite{yoo2017energy}, a generative model was learned for the sequence of user preferences over time. The proposed algorithm can be viewed as a special case of imitation learning. As a future work, authors suggested to use the trust region policy optimization method. The latent distribution of user preferences was learned by an adversarial model in \cite{zhou2019adversarial}, and then, the PG method was used to update the recommender using a set of points of interest sampled according to the learned distribution. The notion of search-session Markov decision process was introduced in \cite{hu2018reinforcement}, which was then used for multi-step ranking. Authors used a version of the PG algorithm to find the optimal policy. A recommender system was proposed in \cite{liu2018deep} that uses the actor-critic algorithm with emphasis on dynamic adaptation.

Here, we provide a comparative study between two categories of deep RL-based recommender systems for online advertising that respectively use value- and policy-based algorithms. As the value-based method, we use different variations of the DQN algorithm and for the policy-based method we use the PG algorithm. For different scenarios and various user/item spaces, the best setting for each method is provided regarding appropriate metrics such as stability of recommendation, training speed, and computational burden. Aiming at designing stable and efficient recommender systems, performance of DQN and PG algorithms are thoroughly studied and compared. The contributions of this paper are three fold:
\begin{itemize}
\item Improving the convergence rate of the recommender system such that the required number of episodes to reach a stable acceptable performance is reduced. 
\item Providing a guaranteed level of performance by improving the performance stability and reducing the variance of the CTR.
\item Providing a computational tool based on the performance area contours to recommend a proper RL algorithm regarding the scenarios that the recommender system is expected to handle.
\end{itemize}

The mentioned improvements are achieved step by step. Two modifications are made in the original DQN architecture. First, instead of using the  convolutional neural network (CNN), an LSTM-based value network is used in the DQN algorithm. Second, the value network is trained using the Huber loss function adopted from robust statistics. The superiority of the modified DQN compared to the original algorithm is shown via simulations for different scenarios with different state/action space dimensions. For the PG algorithm, the policy network is built using the LSTM as well. Both the modified DQN and the PG algorithms owe their improved performance to the gating mechanisms in the LSTM cell that control the flow of information and the ability of LSTM to capture dependencies of clicking on the recommended items to the searching history.     While the DQN-based recommender systems properly handle low dimensional scenarios, the PG-based systems show superiority in handling high dimensional cases.

The rest of the paper is organized as follows. Section~2 reviews the basic concepts and theorems of RL conformed for online advertising. Section~3, presents details on the software testbed used for implementing recommender systems in the RL framework as well as computer experiments and their results. Moreover, guidelines are provided for choosing the best algorithm and setting in different circumstances. Finally, the paper concludes in section~4.

\section{Reinforcment learning}

As previously mentioned, RL refers to learning methods in which an agent interacts with its environment and learns via trial and error. The underlying mathematical model of RL is the Markov decision process (MDP), which is explained in the following.

\subsection{Markov decision process}

Here, the online advertising problem is formulated as an MDP, where the agent is responsible for delivering online advertisements to potential customers. Generally, every MDP consists of a tuple $\left\langle\mathcal{S},\mathcal{A},\mathcal{P},\mathcal{R},\gamma\right\rangle$ with the following components:
\begin{itemize}
\item 
\textbf{State space}, $\mathcal{S}$, refers to the set of all users' profiles viewed as the environment with which the agent (i.e., recommender system) interacts. This set may be partially used by the learning algorithm. Agent can implicitly learn a profile by observing the corresponding user's behaviour over a time interval.
\item 
\textbf{Action space}, $\mathcal{A}$, refers to all available items that can be advertised. If a user watches some items on the e-commerce website for several time intervals, then, as an action $a_t$, the agent recommends an item on the publisher website.
\item 
\textbf{Reward}, $\mathcal{R}$, is defined based on the user's feedback. If the user clicks on the item that the agent recommended, $r_t$ will be 1, otherwise it will be 0.
\item
\textbf{Transition probability}, $\mathcal{P}$, is the probability that the state changes from $s_t$ to $s_{t+1}$. This conditional probability is assumed to satisfy the Markov property. Simply put, this property states that ignore the past and predict the future based on the present: 
\\ $p\left( {{s}_{t+1}}|{{s}_{t}},{{a}_{t}},\ldots ,{{s}_{1}},{{a}_{1}} \right)=p\left( {{s}_{t+1}}|{{s}_{t}},{{a}_{t}} \right) $.
\item
\textbf{Discount factor}, $\gamma$, takes a value between 0 and 1. This value is a measure of the relative importance of future rewards compared to the instant reward. If $\gamma=0$, the agent pays attention only to the immediate reward, on the contrary, if $\gamma=1$, the algorithm considers all the future rewards for taking the current action.
\end{itemize}

\begin{figure}
\begin{center}
\includegraphics[width=1\linewidth]{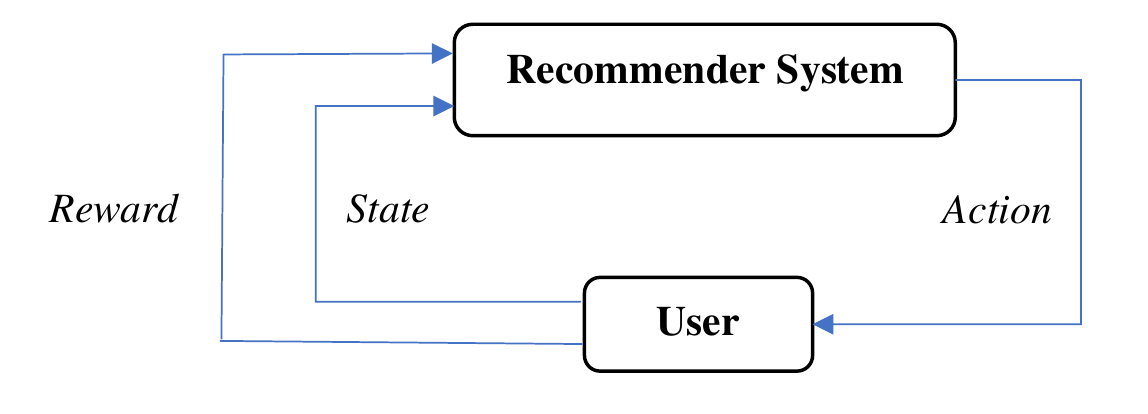}    
\caption{The recommender system as a reinforcement-learning agent.} 
\label{RL_settting}
\end{center}
\end{figure}

Fig. \ref{RL_settting} demonstrates the RL setting. Using the above definitions, the online advertising problem can be formulated as an MDP, which can then  be solved by RL algorithms. Given a history of the corresponding MDP, the goal is to find an optimal policy for advertising $\pi : \mathcal{S} \to \mathcal{A}$, which maximizes the cumulative reward for a specific user. In the context of online advertising, it means that the optimal policy would maximize the click-through rate.

The goal of an agent is to find an optimal policy, which is a policy that maximizes the cumulative rewards at each state, defined as follows:
\begin{equation}
{{V}^{*}}(s)={{\max }_{\pi }}{{\mathrm{E}}^{\pi }}\left\{ \sum\nolimits_{k=0}^{\infty }{{{\gamma }^{k}}{{r}_{t+k}}|{{s}_{t}}=s} \right\}, 
\label{eq1}
\end{equation}
where $\pi$ represents the policy that the agent will follow from state $s$, $\mathrm{E}^{\pi}$ represents the expectation under this policy,  $t$ is the current time step, and $t+k$ refers to future time steps. Furthermore, $r_{t+k}$ is the immediate reward at time step $(t+k)$. The problem will be solved, if the recommender agent finds the optimal state-action value function, which is defined for all situations and actions as:
\begin{equation}
{{Q}^{*}}(s,a)={{\max }_{\pi }}{{\mathrm{E}}^{\pi }}\left\{ \sum\nolimits_{k=0}^{\infty }{{{\gamma }^{k}}{{r}_{t+k}}|{{s}_{t}}=s,a_t=a} \right\} 
\label{eq2}
\end{equation}
Considering a finite control horizon, $T$, the objective function can be rewritten as follows:
\begin{equation}
{\mathrm{E}^{\pi }}\left\{ \sum\nolimits_{k=0}^{T}{{{\gamma }^{k}}{{r}_{t+k}}|{{s}_{t}}=s} \right\} 
\label{eq3}
\end{equation}

An optimal policy can be found from either the optimal state value function, ${V}^{*}(s)$, or the optimal state-action value function, ${Q}^{*}(s,a)$. For problems with finite small state and action spaces, methods such as dynamic programming and temporal-difference (TD) learning can be implemented using a table to store and update the value function. However, function approximation methods are widely used when it comes to high-dimensional problems in order to cope with the curse of dimensionality. These methods can benefit from parametric models such as deep neural networks to represent the value and policy functions. Then, the algorithm will optimize the model parameters based on the reward signal. As two successful deep RL algorithms that rely on deep learning for function approximation, we can refer to DQN and PG, which have been used for designing recommender systems.

\subsection{Deep Q-Network}

DQN was originally used for playing Atari games \cite{mnih2013playing}. In DQN, convolutional neural networks are used to approximate the state-action value function and stochastic gradient descent is used to optimize the objective function and update the parameter values. The experience replay mechanism is also adopted to address the issues of interrelated data and unstable distribution. The optimal state-action value function is the expected value of $r + \gamma Q^{*}(s^{'} , a^{'})$. 
The Q-network is trained by minimizing the mean squared error (MSE) loss function:
\begin{equation}
{{L}_{i}}({{\theta }_{i}})={{E}_{s,a \sim \rho (.)}}\left[ {{({{y}_{i}}-Q(s,a;{{\theta }_{i}}))}^{2}} \right]
\label{eq3}
\end{equation}
Algorithm 1 presents the pseudo code of the DQN algorithm with experience replay \cite{mnih2013playing}.

\begin{algorithm*}[!h]

    \begin{algorithmic}[1]

        \caption{Deep Q-learning with Experience Replay \cite{mnih2013playing}} \label{algorithm: cds bw}
        \State Initialize the replay memory D to capacity N
        \State Initialize the state-action value function Q with random weights
        \ForAll{$episode = 1:M$}
        		\State Initialize the sequence ${s_1}={x_1}$ and the preprocessed       		
        		sequence 
        		  $\phi_1=\phi(s_1)$		 
            \ForAll {$t = 1:T$}
            \State With probability $\epsilon$ select a random action $a_t$,
            \State Otherwise select $ {{a}_{t}}={{\max }_{a}}{{Q}^{*}}\left( \phi \left( {{S}_{t}} \right),a;					  \theta \right)$
            \State Execute the action $a_t$ in the emulator and observe the reward $r_t$ and image $x_{t+1}$
            \State Set $s_{t+1}=s_{t},a_{t},x_{t+1}$ and preprocess $\phi_{t+1}=\phi(s_{t+1})$
 						\State Store the transition $(\phi_{t},a_{t},r_{t},\phi_{t+1})$ in $\mathcal{D}$
            \State Sample the random minibatch of transitions $(\phi_{j},a_{j},r_{j}\phi_{j+1})$ from $\mathcal{D}$
             \State Set ${{y}_{i}}=$                    $\left\{\begin{matrix} {{r}_{j}} & \text{for terminal }{{\phi }_{j+1}}  \\ 
{{r}_{j}}+\gamma {{\max }_{{{a}'}}}Q\left( {{\phi }_{j+1}},{a}';\theta  \right) & \text{for non-terminal }{{\phi }_{j+1}}  \\ 
									\end{matrix}\right. $
						\State Perform a gradient descent step on ${{({{y}_{j}}-Q({{\phi }_{j}},{{a}_{j}};\theta ))}^{2}}$ 
        \EndFor
    \EndFor             
    \end{algorithmic}
\end{algorithm*}

\subsection{Policy Gradient}

For the PG method, the decision-making process is carried out by the agent $\forall s\in \mathcal{S}, a \in \mathcal{A}$ at any time step  according to a parameterized policy: 
\begin{equation}
\pi(s,a,\theta)=Pr\{a_t=a|s_t=s, \theta\}
\label{eq3}
\end{equation}
where $\theta \in \mathcal{R}^l$ for $l<<|\mathcal{S}|$ is an array of parameters used to approximate the optimal policy function by a deep neural network. It is assumed that $\pi$ is differentiable with respect to parameters, so that it can be optimized using gradient-based methods. Two formulations of this problem were presented in \cite{sutton2000policy}: the average-reward and the start-state formulation, which are reviewed in what follows.

The \textit{average-reward} formulation prioritizes the policies based on their long-term expected collected reward per step:
\begin{equation}
\begin{aligned} 
 \rho (\pi ) &= \underset{n\to \infty }{\mathop{\lim }}\,\frac{1}{n}E\left\{ {{r}_{1}}+{{r}_{2}}+\cdots +{{r}_{n}}|\pi  \right\} \\  &=
   \sum\limits_{s}{{{d}^{\pi }}(s)\sum\limits_{a}{\pi (s,a)\mathcal{R}_{s}^{a}}}  \\ 
  {{Q}^{\pi }}(s,a) &= \sum\limits_{t=1}^{\infty } E\left\{ {r}_{t} - \rho(\pi)|{{s}_{0}}=s,{{a}_{0}}=a,\pi  \right\}, \\ & \forall s \in \mathcal{S}, a \in \mathcal{A} 
\end{aligned} 
\label{eq4}
\end{equation}
where 
\begin{equation}
{{d}^{\pi }}(s)={{\lim }_{t\to \infty }}\Pr \left\{ {{s}_{t}}=s|{{s}_{0}},\pi  \right\}
\end{equation}
denotes a stationary distribution over states given the policy, which is assumed to exist for all policies regardless of the initial state, $s_0$. 

The \textit{start-state} formulation focuses on long-term reward collected from a specific initial state, $s_0$, which is formulated as:
\begin{equation}
\begin{aligned} 
 & \rho (\pi )=E\left\{ \sum\limits_{t=1}^{\infty }{{{\gamma }^{t-1}}{{r}_{t}}\left| {{s}_{0}},\pi  \right.} \right\} \\ 
 & {{Q}^{\pi }}(s,a)=E\left\{ \sum\limits_{k=1}^{\infty }{{{\gamma }^{k-1}}{{r}_{t+k}}|{{s}_{t}}=s,{{a}_{t}}=a,\pi } \right\} \\ 
\end{aligned} 
\label{eq5}
\end{equation}
where $\gamma \in [0,1]$ is the discount factor and $d^{\pi}(s)$ is considered as a discounted weighting of states reachable from $s_0$ following policy $\pi$:
\begin{equation}
\text{  }{{d}^{\pi }}(s)=\sum\limits_{t=0}^{\theta }{{{\gamma }^{t}}\Pr \left\{ {{s}_{t}}=s|{{s}_{0}},\pi  \right\}}
\label{eq6}
\end{equation}
To optimize the policy network parameters, gradient of the performance metric with respect to parameters is needed, which is calculated for both formulations as \cite{sutton2000policy}:  
\begin{equation}
\frac{\partial \rho }{\partial \theta }=\sum\limits_{s}{{{d}^{\pi }}(s})\sum\limits_{a}{\frac{\partial \pi (s,a)}{\partial \theta }}{{Q}^{\pi }}(s,a) 
\label{eq7}
\end{equation}
After training, the policy network can be fed with a description of state to receive a distribution of actions at the output \cite{mousavi2017traffic}.

\section{Experiments}

Since RL algorithms work in an online manner, it is necessary to use an environment that is capable of handling random appearance of users, ads, and suggested items for testing and evaluating such algorithms. Due to this fact, we adopted the RecoGym environment \cite{rohde2018recogym}, which provides the settings required for an RL problem. It allows for interaction between an agent and the environment, which leads to receiving a reward from the environment, when a user clicks on a recommended item.

\begin{figure*}
\begin{center}
\includegraphics[width=0.55\linewidth]{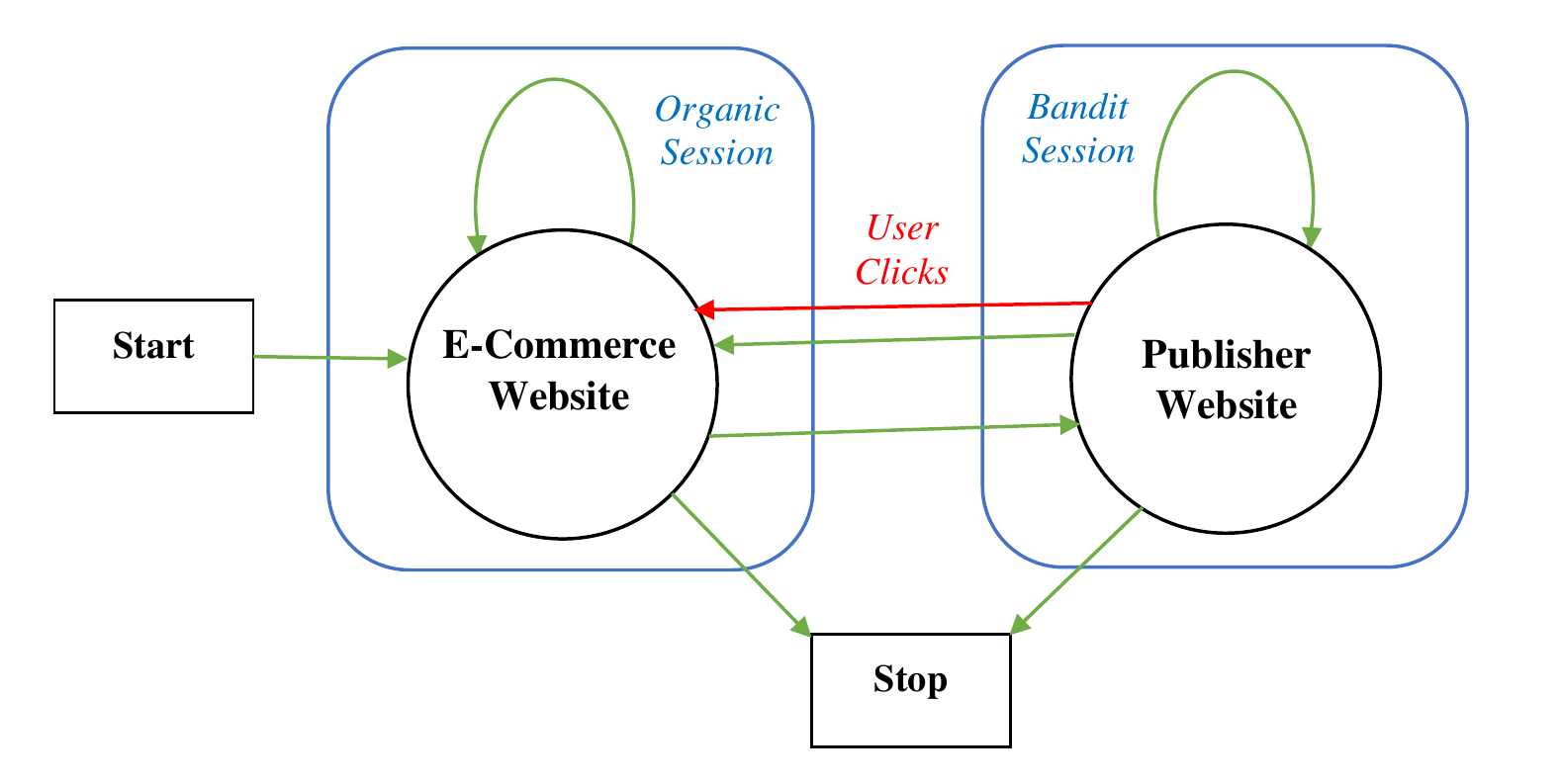}   
\caption{Markov chain of the organic and bandit sessions in the RecoGym environment.} 
\label{MDP}
\end{center}
\end{figure*}

The RecoGym environment is mainly designed for online advertising, and the underlying process has two parts: 
\begin{itemize}
\item The \textit{organic} session occurs on the e-commerce website during which a user sees various items. 
\item The \textit{bandit} session occurs on the publisher website, where the agent has the opportunity to recommend some items to users and observe their reactions.
\end{itemize}
The Markov chain of this environment is depicted in Fig. \ref{MDP}, which shows how these two parts are related  \cite{rohde2018recogym}. First, a user enters the organic environment and sees different items at different time steps. Thus, within a variable time interval $T$, a user observes one or more items. Then, the bandit session starts, and based on the user's search history at previous time steps, the recommender agent suggests several items to him/her. The Bandit session ends after a time interval, which is randomly chosen for different people. The user may click on one of the recommended products during this session. If this happens, the recommender agent receives a reward and the user will move back to the organic session. Otherwise, transfer to the organic session will occur when the bandit  session ends. The switching between these two sessions occurs for a random number of times for each user.

To train the recommender agent, we first deploy the DQN algorithm \cite{mnih2013playing} using a one-dimensional CNN to approximate the state-action value function. Since the number of observations and the length of time intervals change, variable padding is used for observations before feeding them to the convolutional network. Exploration rate decreases from 0.9 to 0.1 during each episode. We assume that each user's profile contains ten features. Two different loss functions are considered: the MSE and the Huber function in order to demonstrate effect of loss on the performance. The Huber function, which has been used in robust regression, has very low sensitivity to outliers compared to  the mean-squares error. The Huber function is defined as follows: 
\begin{equation}
{{L}_{\delta }}\left( y,f(x) \right)=\left\{ \begin{matrix} 
   \frac{1}{2}{{\left( y-f(x) \right)}^{2}} & for\left| y-f(x) \right|\le \delta   \\ 
   \delta \left| y-f(x) \right|-\frac{1}{2}{{\delta }^{2}} & otherwise  \\ 
\end{matrix} \right .
\label{eq3}
\end{equation}

\begin{figure}[!ht]
\centering
\begin{subfigure}[b]{0.4\textwidth}
\includegraphics[width=\textwidth]{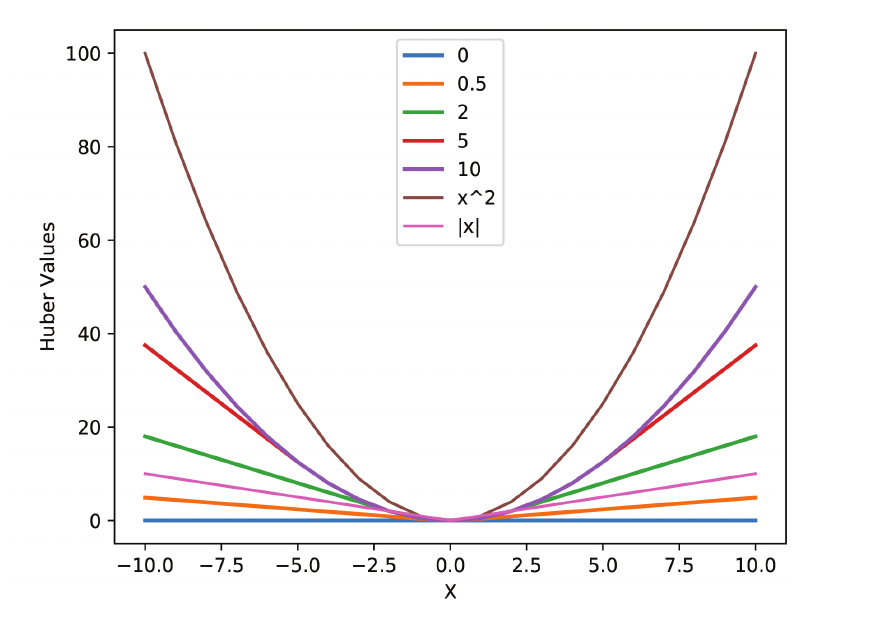}
\caption{}
\label{Huber}
\end{subfigure}
\quad
\begin{subfigure}[b]{0.38\textwidth}
\includegraphics[width=\textwidth]{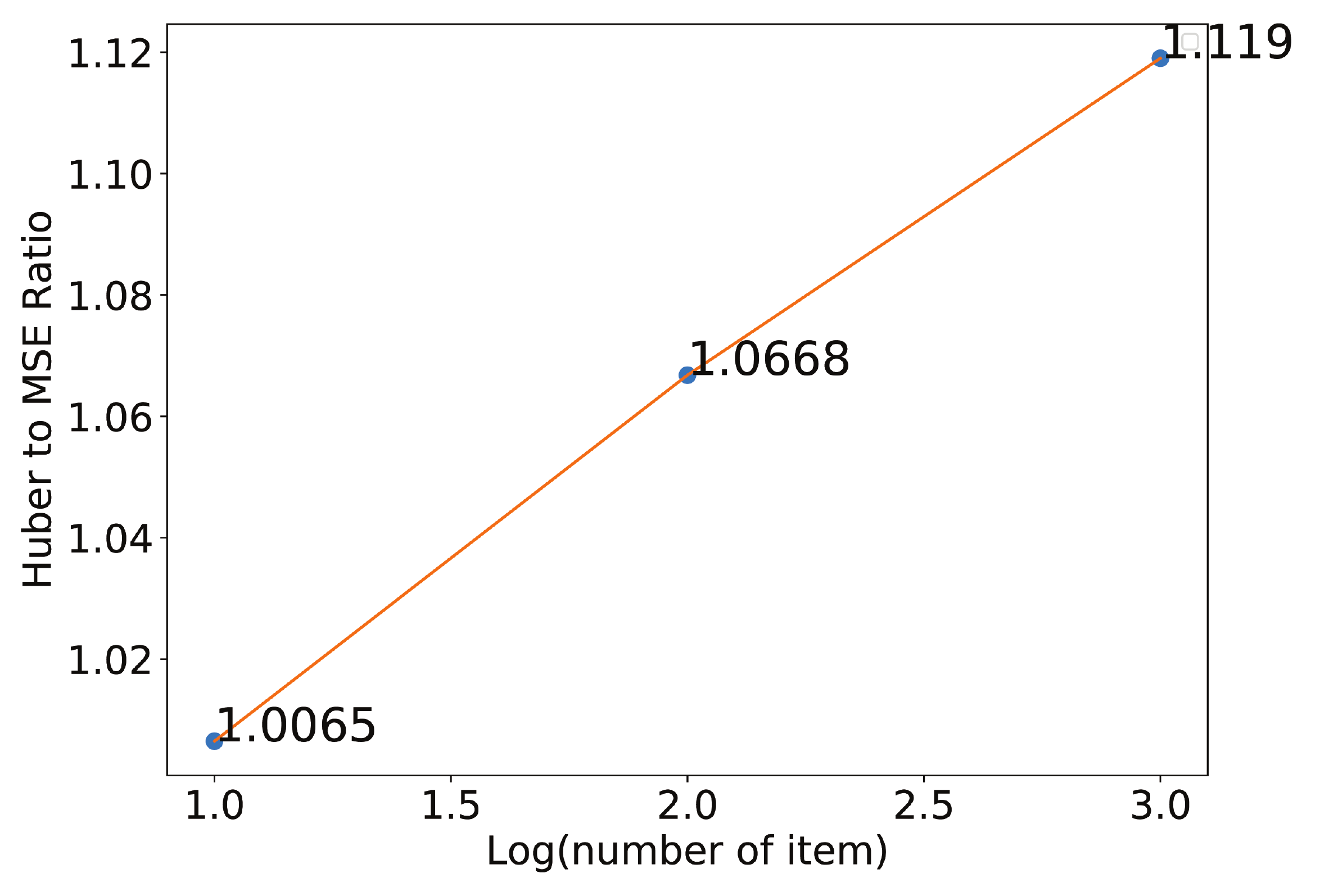}
\caption{}
\label{huber_eval_1}
\end{subfigure}
\caption{(a) Huber loss function for different values of $\delta$ compared to $x^2$ and $|x|$. (b) Average performance ratio of the DQN algorithm using Huber and MSE loss functions for different number of items.}\label{Huber_loss}
\end{figure}

\begin{figure}[!ht]
\centering
\begin{subfigure}[b]{0.45\textwidth}
\includegraphics[width=\textwidth]{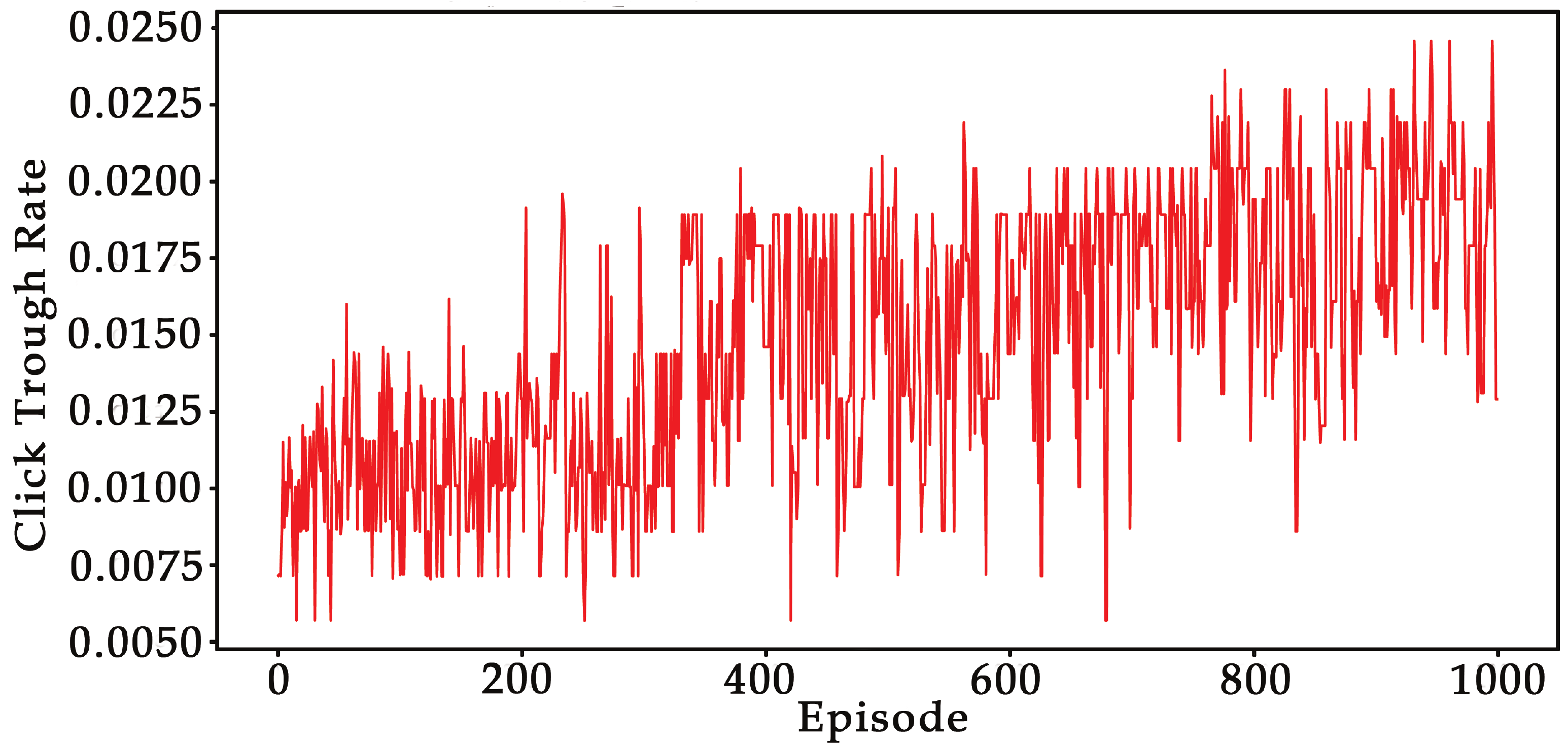}
\caption{}
\label{cnn100}
\end{subfigure}
\quad
\begin{subfigure}[b]{0.45\textwidth}
\includegraphics[width=\textwidth]{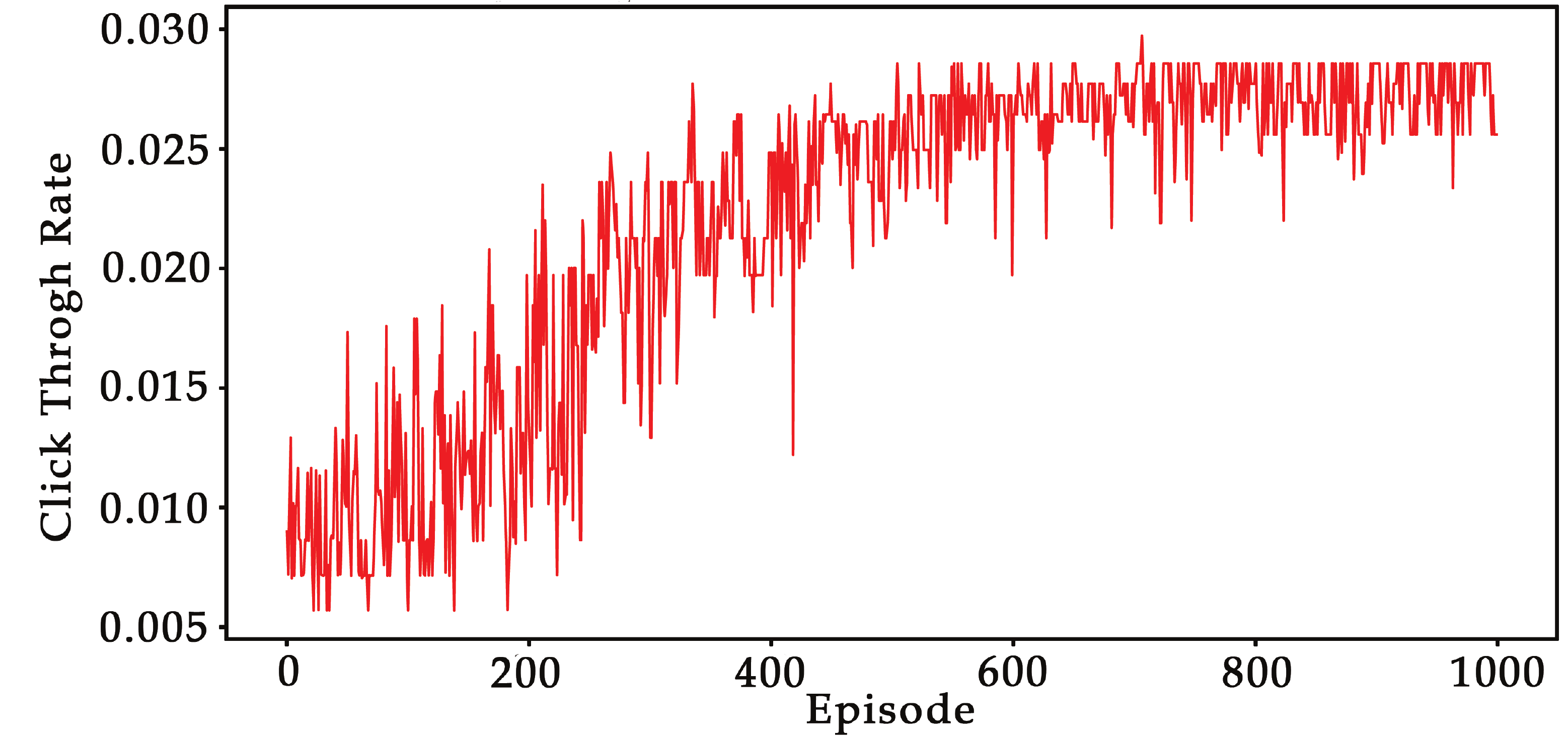}
\caption{}
\label{lstm100}
\end{subfigure}
\caption{Click-through rate versus the number of  episodes for 100 users and 100 items achieved by: (a) DQN with CNN and (b) DQN with LSTM}\label{cnn_lstm_1}
\end{figure}

\begin{figure}[!ht]
\centering
\begin{subfigure}[b]{0.45\textwidth}
\includegraphics[width=\textwidth]{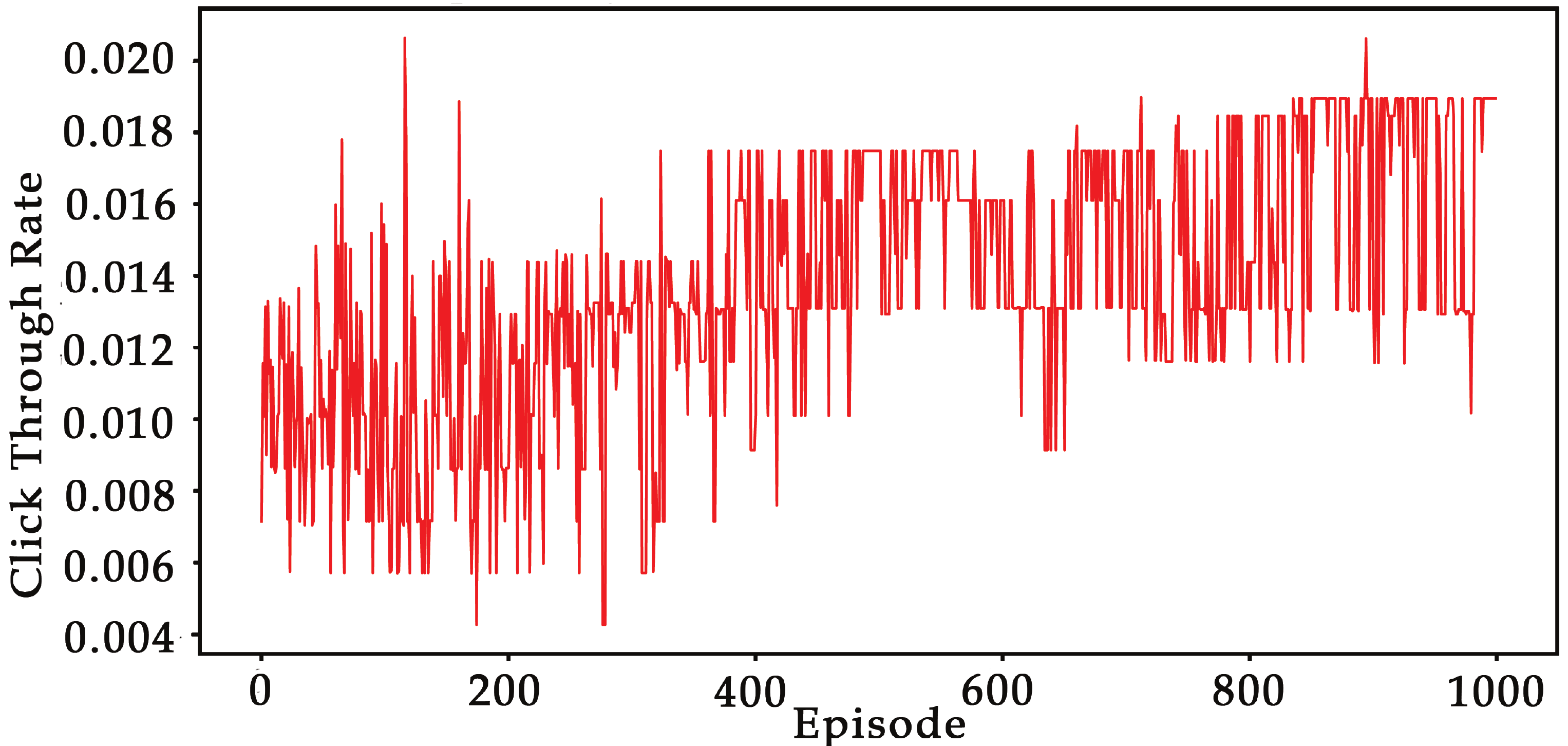}
\caption{}
\label{cnn1000}
\end{subfigure}
\quad
\begin{subfigure}[b]{0.45\textwidth}
\includegraphics[width=\textwidth]{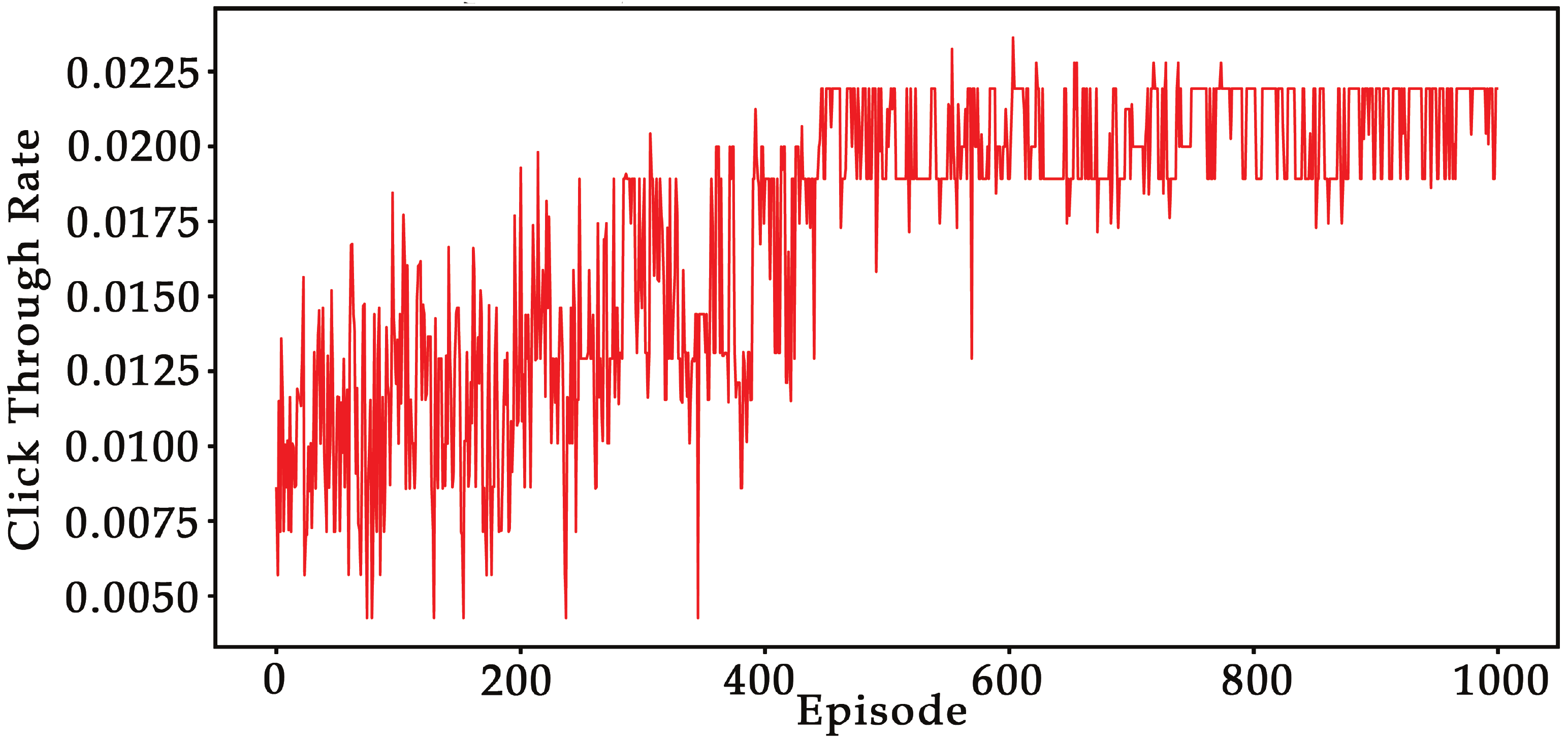}
\caption{}
\label{lstm1000}
\end{subfigure}
\caption{Click-through rate versus the number of  episodes for 1000 users and 1000 items achieved by: (a) DQN with CNN and (b) DQN with LSTM}\label{cnn_lstm_2}
\end{figure}

\begin{figure}[!ht]
\centering
\begin{subfigure}[b]{0.45\textwidth}
\includegraphics[width=\textwidth]{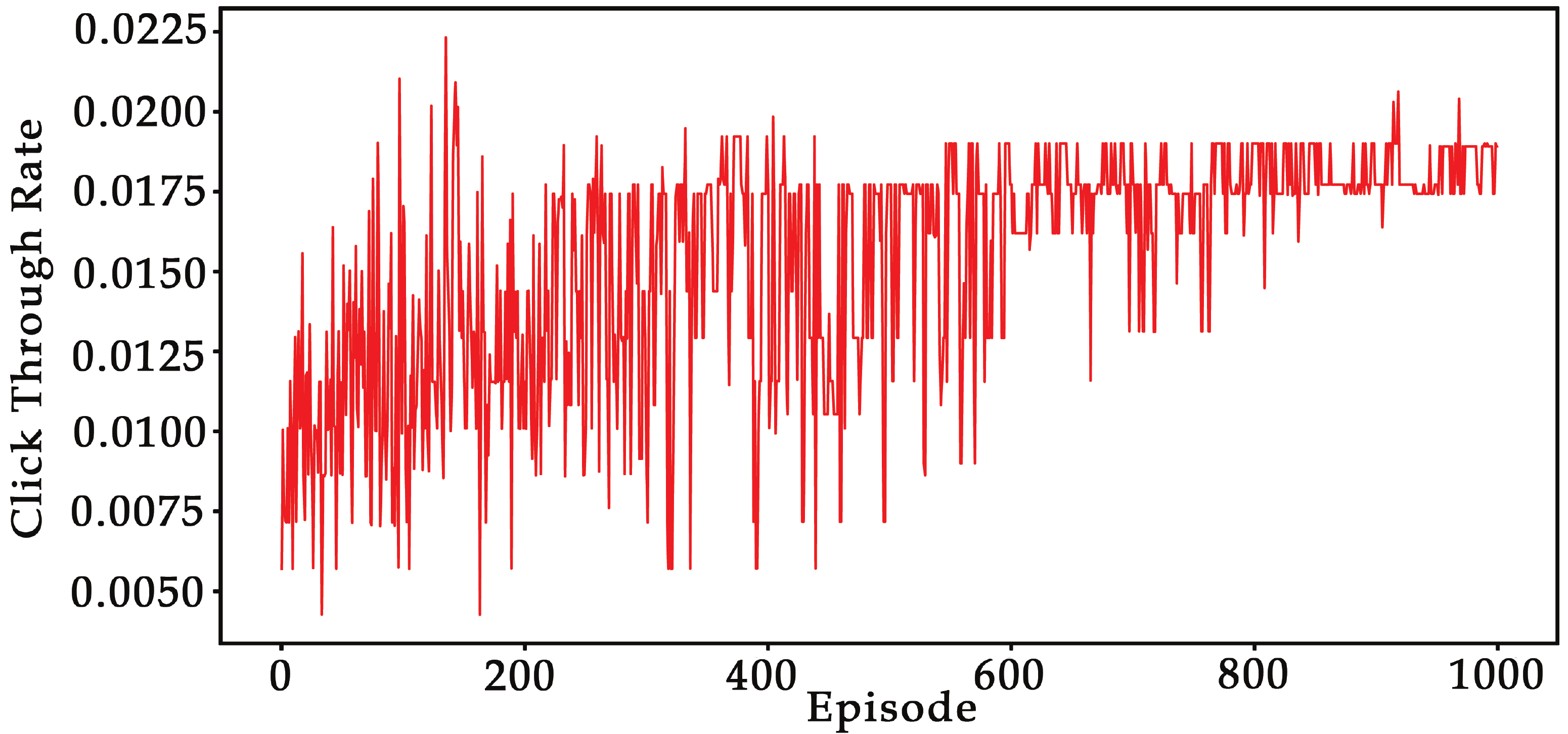}
\caption{}
\label{cnn10000}
\end{subfigure}
\quad
\begin{subfigure}[b]{0.45\textwidth}
\includegraphics[width=\textwidth]{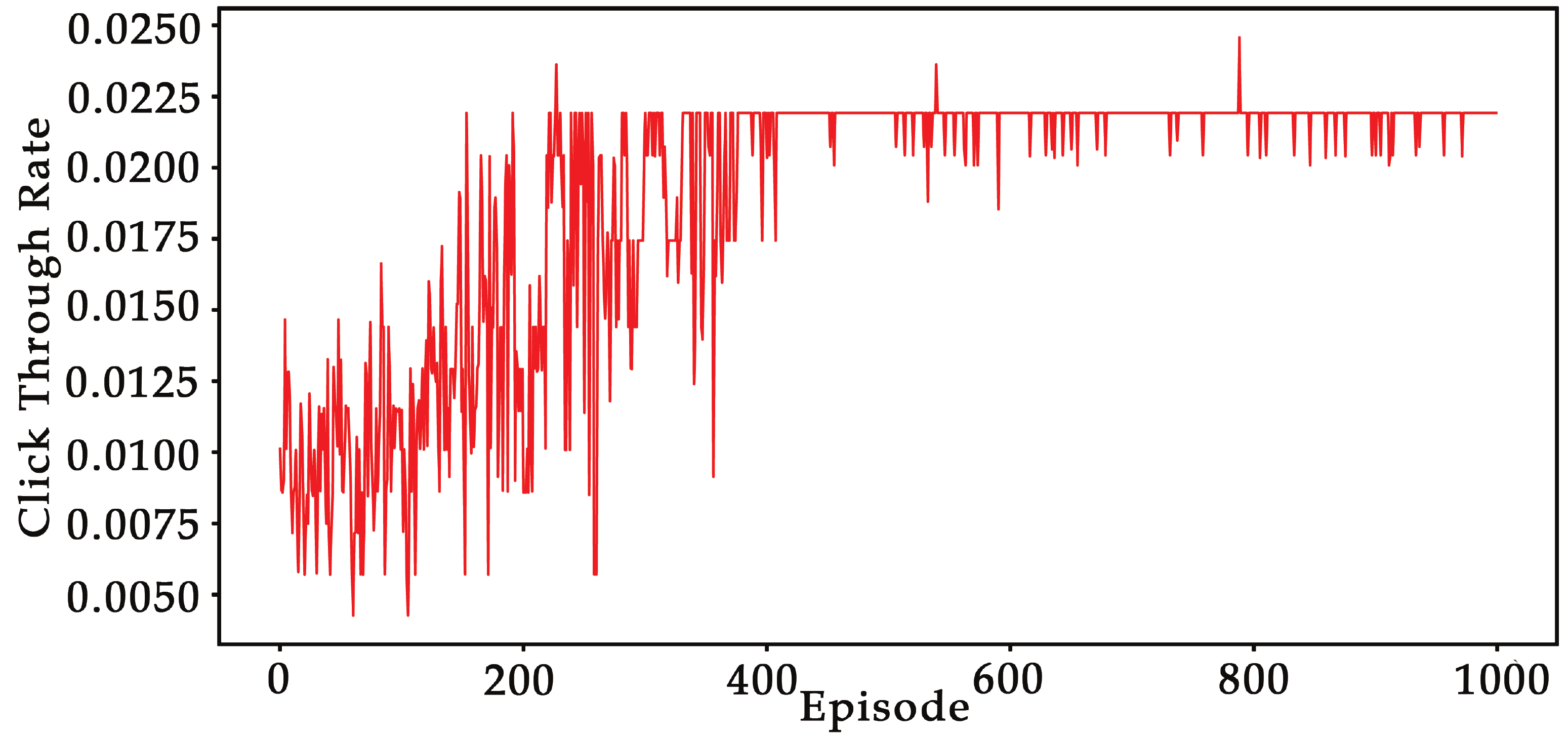}
\caption{}
\label{lstm10000}
\end{subfigure}
\caption{Click-through rate versus the number of  episodes for 10000 users and 10000 items achieved by: (a) DQN with CNN and (b) DQN with LSTM}\label{cnn_lstm_3}
\end{figure}

Figure \ref{Huber} illustrates the Huber loss function for different values of $\delta$ and compares it with $x^2$ and $|x|$. In all experiments, we set $\delta=2$. The recommender agent was built around the DQN algorithm using both  Huber and MSE loss functions. Results reported here were obtained by averaging over 50 runs. Average performance ratio of the DQN algorithm for these two loss functions is depicted in Fig. \ref{huber_eval_1} for 10, 100, and 1000 items. Compared to MSE, using the Huber function shows an increase of 12 percent in the CTR as the number of items increases from 10 to 1000. Figures \ref{cnn100}, \ref{cnn1000}, and \ref{cnn10000} show the CTR achieved by the DQN algorithm using the Huber loss function versus the number of episodes for different state-action spaces (i.e., 100, 1000, and 10000 users and items).

Looking for different items on an e-commerce website by a user can be viewed as a sequential process in which order and history are of crucial importance. Item observation by a user in each time step depends on previously observed items during past time steps. Hence, we can expect that deploying recurrent neural networks would provide a better approximation of the state-action value function. We used a modified version of the DQN algorithm by replacing the convolutional neural network with the LSTM, which is a recurrent neural network that benefits from gating mechanisms for controlling the flow of information. The LSTM is used to approximate the state-action value function in the DQN algorithm. We repeated the previously mentioned experiments with this new architecture of the DQN algorithm that uses both the LSTM and Huber loss function. Again, the number of each user's features is assumed to be ten, and exploration rate decreases from 0.9 to 0.1 in each episode. Results are shown in figures \ref{lstm100}, \ref{lstm1000}, and \ref{lstm10000} for 100, 1000, and 10000 users and items, respectively. In these figures, superiority of the new setting that deploys LSTM instead of CNN is obvious in both convergence and performance. Moreover, for larger action spaces (i.e., more items), the LSTM-based DQN shows improved stability and lower variance compared to the CNN-based DQN. 

\begin{figure}[!ht]
\centering
\begin{subfigure}[b]{0.45\textwidth}
\includegraphics[width=\textwidth]{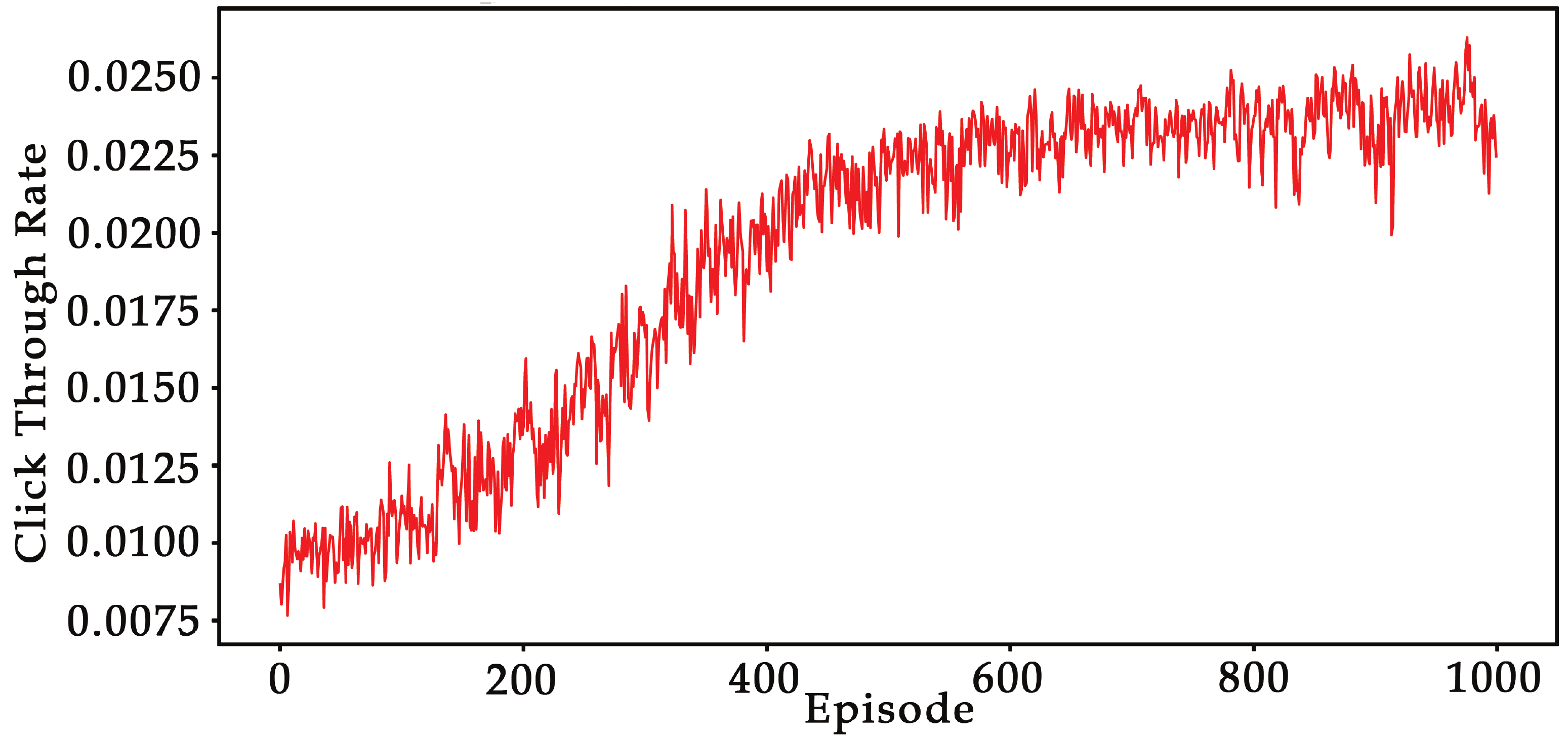}
\caption{}
\label{}
\end{subfigure}
\quad
\begin{subfigure}[b]{0.45\textwidth}
\includegraphics[width=\textwidth]{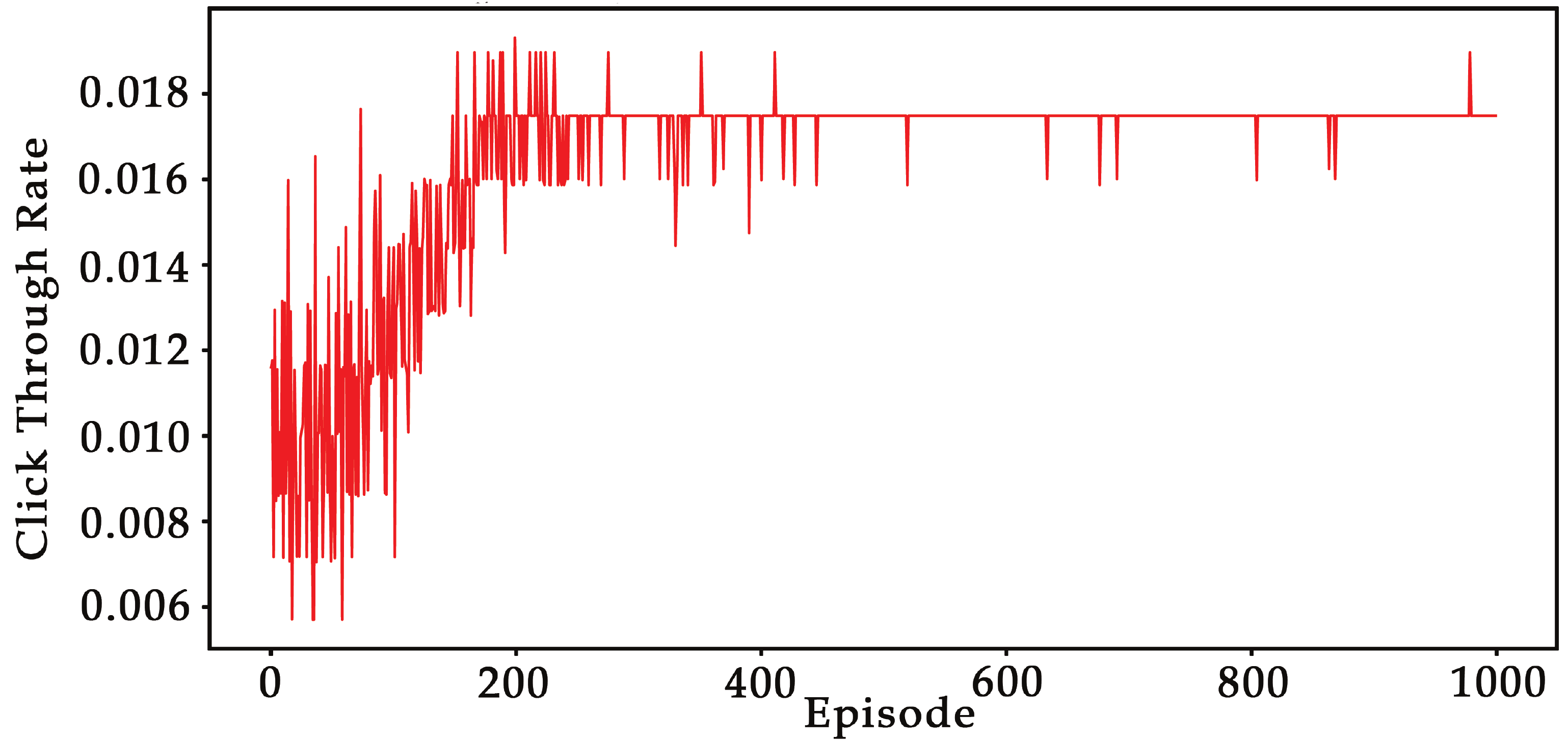}
\caption{}
\label{huber_eval}
\end{subfigure}
\caption{Click-through rate versus the number of  episodes for 100 users and 100 items achieved by: (a) DQN and (b) PG}\label{dqn_pg_1}
\end{figure}

\begin{figure}[!ht]
\centering
\begin{subfigure}[b]{0.45\textwidth}
\includegraphics[width=\textwidth]{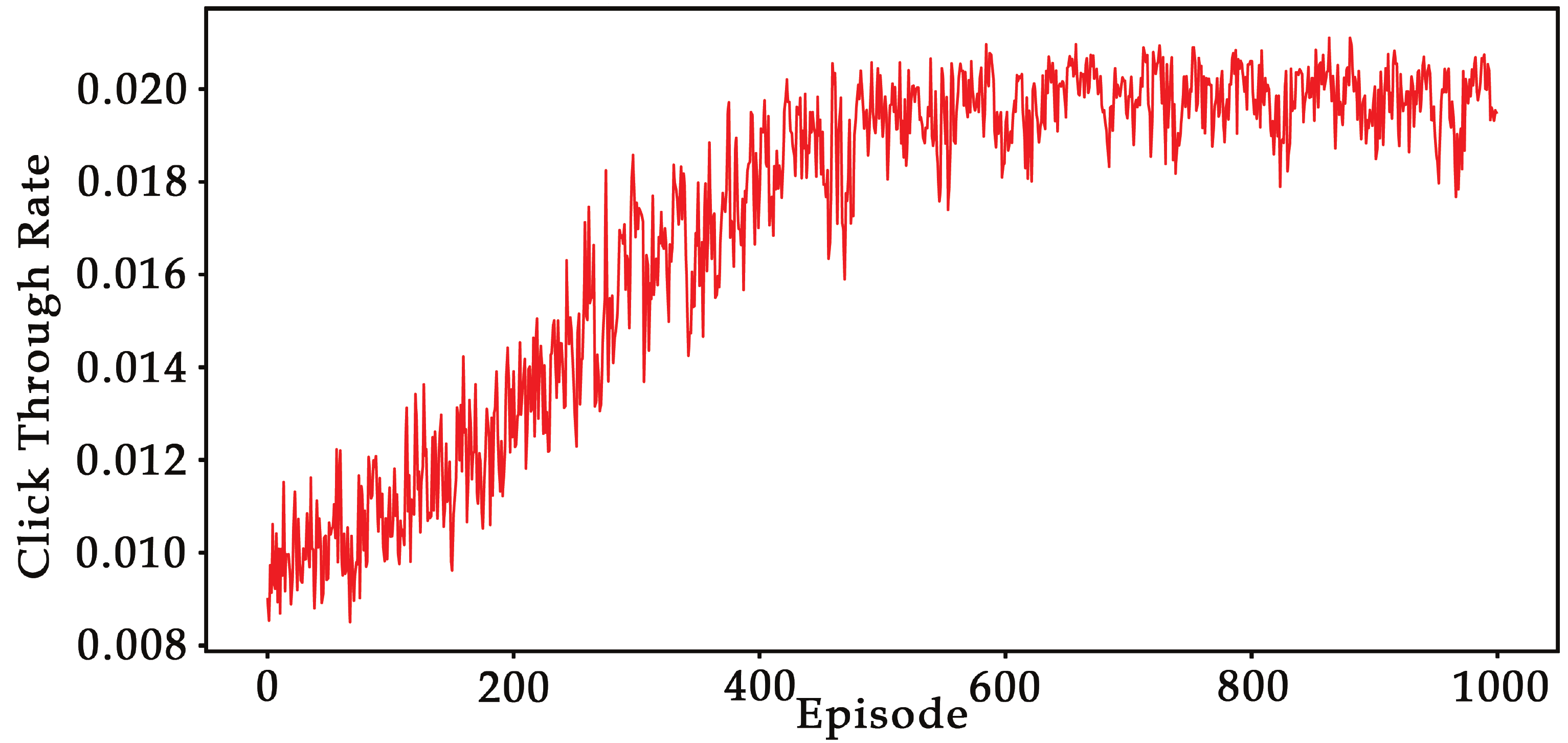}
\caption{}
\label{}
\end{subfigure}
\quad
\begin{subfigure}[b]{0.45\textwidth}
\includegraphics[width=\textwidth]{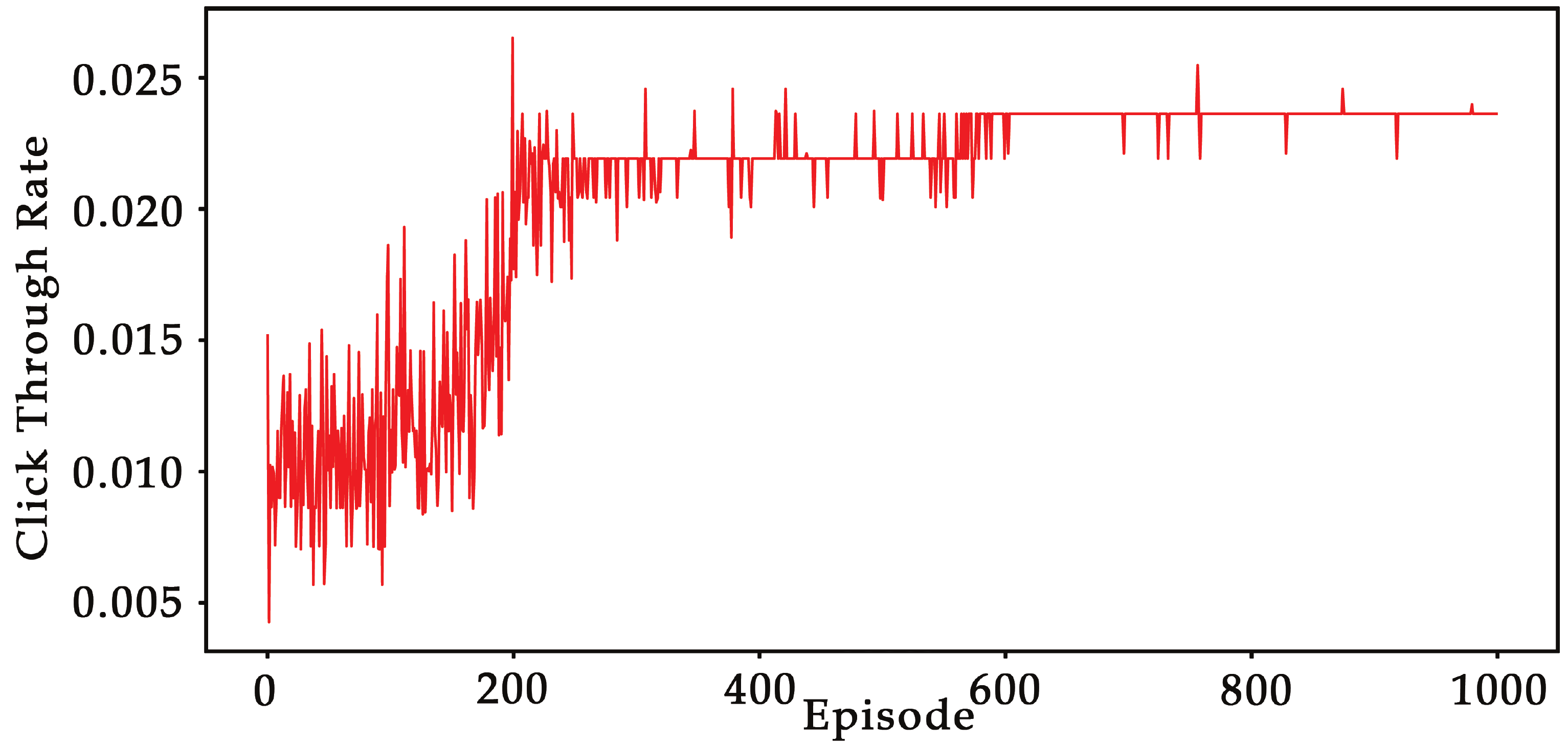}
\caption{}
\label{huber_eval}
\end{subfigure}
\caption{Click-through rate versus the number of  episodes for 1000 users and 1000 items achieved by: (a) DQN and (b) PG}\label{dqn_pg_2}
\end{figure}

\begin{figure}[!ht]
\centering
\begin{subfigure}[b]{0.45\textwidth}
\includegraphics[width=\textwidth]{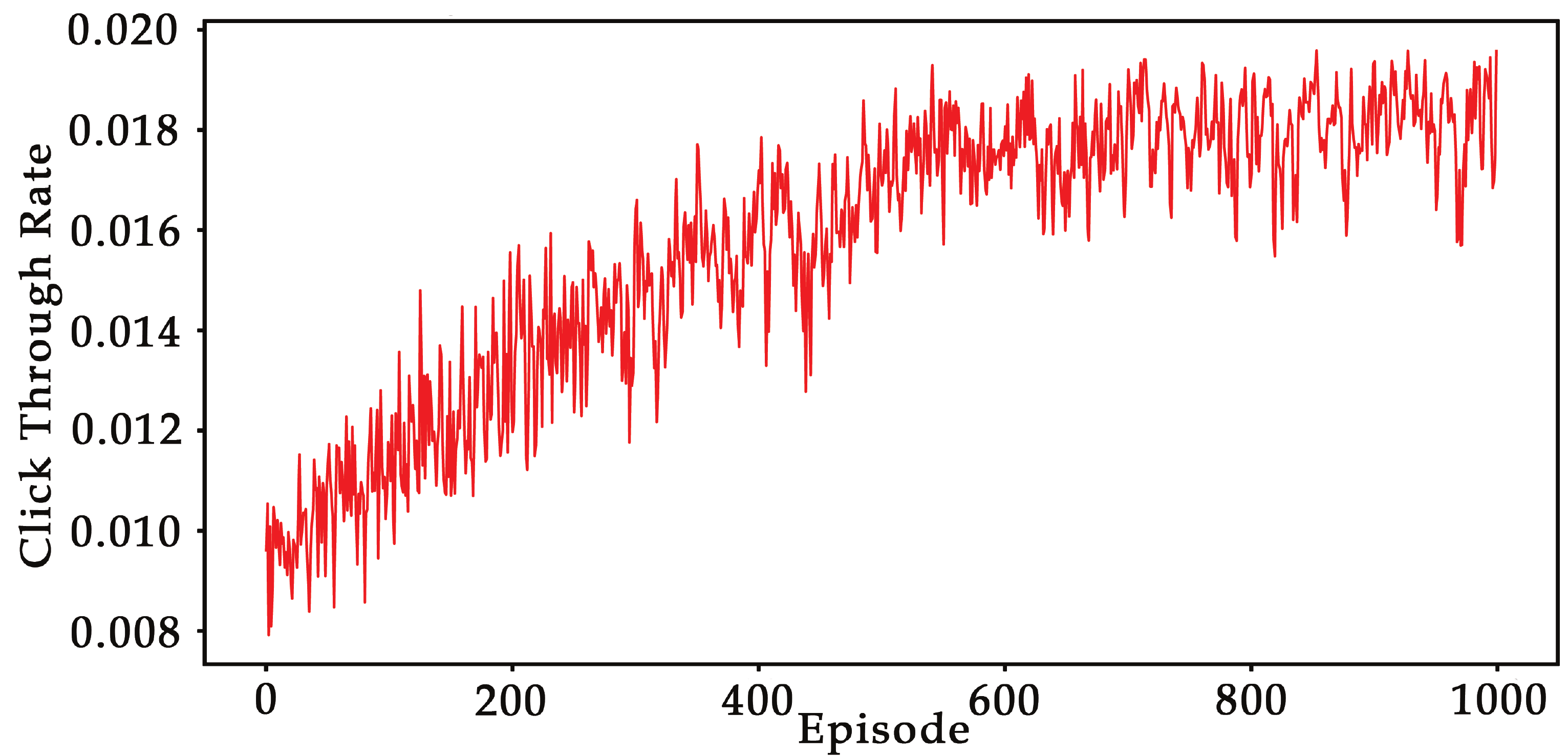}
\caption{}
\label{}
\end{subfigure}
\quad
\begin{subfigure}[b]{0.45\textwidth}
\includegraphics[width=\textwidth]{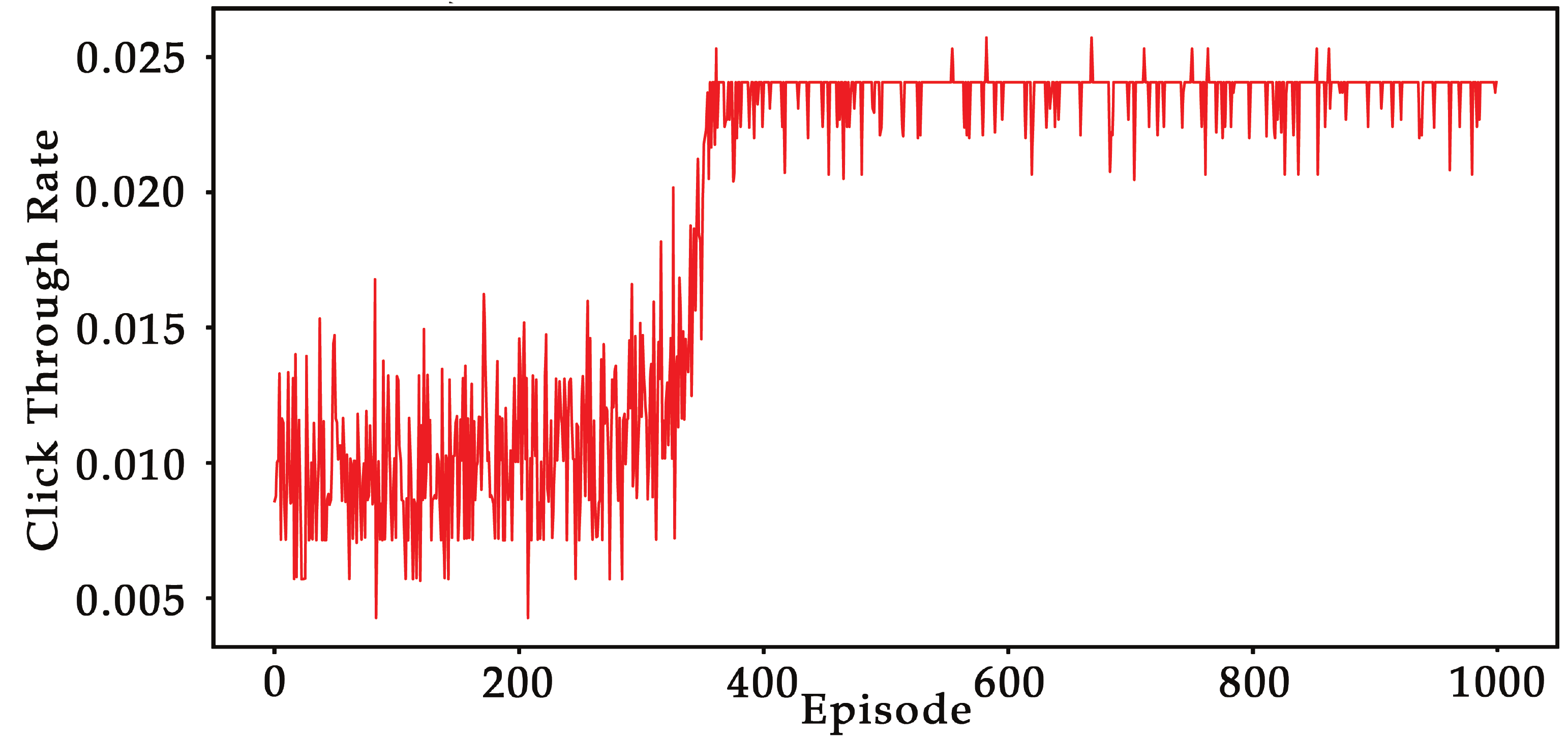}
\caption{}
\label{huber_eval}
\end{subfigure}

\caption{Click-through rate versus the number of  episodes for 10000 users and 10000 items achieved by:  (a) DQN and (b) PG}\label{dqn_pg_3}
\end{figure}

Next, we trained a recommender system based on the policy gradient method using the RecoGym environment. While the DQN learns an approximation of the optimal state-action value function and from that finds the optimal policy, the PG algorithm directly finds the optimal policy by searching the policy space. The PG algorithm was implemented using a policy network that learns a distribution over the actions. Hidden layers of the policy network included LSTM units and dense layers, and the output layer used the softmax function to form a probability distribution. The categorical cross-entropy loss function was used for training the policy network. Results achieved by the PG algorithm are compared with those of the DQN in figures \ref{dqn_pg_1}, \ref{dqn_pg_2}, and \ref{dqn_pg_3} for 100, 1000, and 10000 users and items, respectively. From these figures, we see that the PG method converges faster than the DQN, which is more significant in scenarios with larger state/action spaces. Moreover, The PG algorithm achieves a better CTR with less fluctuations and lower variances, especially for larger state/action spaces, and in effect therefore, presents a more stable behaviour.

\begin{figure}[!ht]
\centering
\begin{subfigure}[b]{0.45\textwidth}
\includegraphics[width=\textwidth]{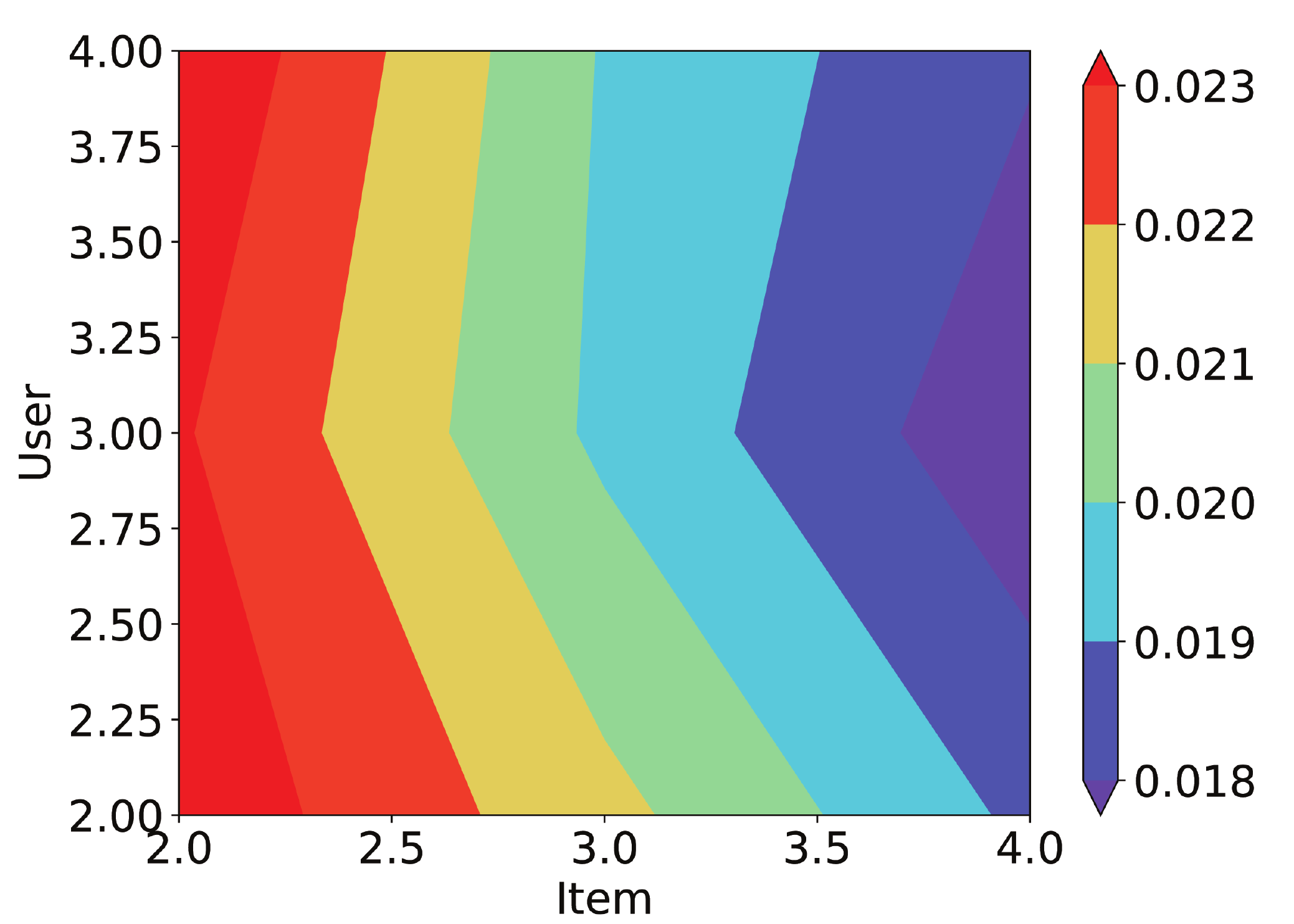}
\caption{DQN}
\label{dqn_area}
\end{subfigure}
\quad
\begin{subfigure}[b]{0.45\textwidth}
\includegraphics[width=\textwidth]{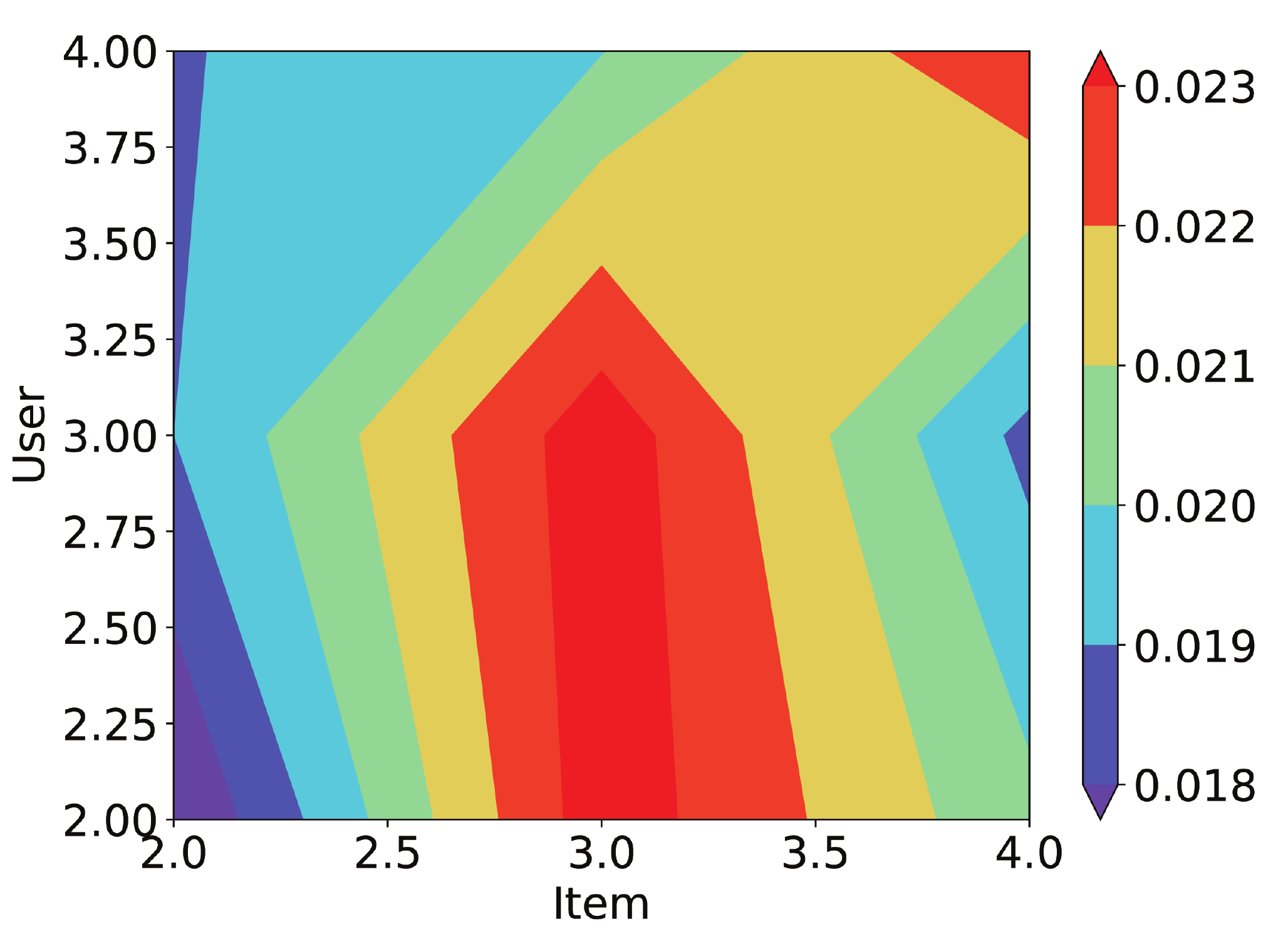}
\caption{PG}
\label{pg_area}
\end{subfigure}
\caption{Performance area for different RL algorithms: (a) DQN and (b) PG. Both axes are in logarithmic scale.}\label{}
\end{figure}

In order to obtain a measure that guides us in choosing a proper deep RL algorithm for designing a recommender system for a specific scenario, we need to examine the user-item space. Figures \ref{dqn_area} and \ref{pg_area} illustrate the performance of the DQN and the PG algorithms for state/action spaces with different dimensions (i.e., different number of users and items). These figures were plotted using logarithmic-scale for both axes. According to these figures, the DQN and the PG algorithms can be respectively recommended for lower and higher dimensional state/action spaces.

\section{Conclusion}

We deployed two deep reinforcement learning algorithms to design recommender systems for online advertising: the Deep Q-network and the policy gradient method. While the former is a critic-only algorithm, the latter is an actor-only one. The RecoGym was used as the environment with which the RL agent interacts. The RL agent aims at maximising the click-through rate for the recommended items. The original architecture of the DQN was modified using the LSTM as the value network to take account of the key role that order plays in the search history and observing different items by users. Moreover, the Huber function was adopted from robust statistics as the loss function. These two modifications improved the convergence characteristics of the DQN algorithm and led to less fluctuations and lower variance in the click-through rate. Regarding the importance of order in the sequence of observations of different items by users, the LSTM was used to implement the policy network for the PG algorithm as well. The PG algorithm showed even a better convergence behaviour and lower variance in the click-through rate compared to the LSTM-based DQN, especially for larger state/action spaces. These results confirm the fact that actor-only algorithms are more resilient against fast-changing and nonstationary environments compared to critic-only algorithms. Finally, performance area contours were used to provide guidelines for choosing a proper deep reinforcement learning algorithm for building recommender systems based on dimensions of the state/action spaces (i.e., the number of users and items that the recommender system is expected to handle).

\bibliographystyle{IEEEtran}
\bibliography{IEEEabrv,Main}

\end{document}